\documentclass[journal]{IEEEtran}
\usepackage{amsmath,amsfonts,amssymb}
\usepackage{algorithmic}
\usepackage{array}
\usepackage[caption=false,font=normalsize,labelfont=sf,textfont=sf]{subfig}
\usepackage{textcomp}
\usepackage{color}
\usepackage{stfloats}
\usepackage{url}
\usepackage{verbatim}
\usepackage{graphicx}
\usepackage{mathtools}
\usepackage{amsthm}
\usepackage{soul}
\usepackage{pifont}
\newcommand\xy{\textcolor{black}}

\usepackage[table]{xcolor}
\definecolor{oursgray}{gray}{0.90}

\usepackage{multirow}
\usepackage{hyperref}
\usepackage{url}
\usepackage{makecell}
 \usepackage{multirow,booktabs}
\usepackage{graphicx}
\usepackage{enumitem}
\usepackage{wrapfig}
\begin{document}
\title{PMSN: A Parallel Multi-compartment Spiking Neuron for Multiscale Temporal Processing}

\author{Xinyi Chen, Jibin~Wu,~\IEEEmembership{Member,~IEEE}, Chenxiang~Ma, Yinsong~Yan, Hanwen~Liu, Yujie~Wu, Kay~Chen~Tan,~\IEEEmembership{Fellow,~IEEE}

\IEEEcompsocitemizethanks{
\IEEEcompsocthanksitem This work is accepted by IEEE Transactions on Neural Networks and Learning Systems, DOI:10.1109/TNNLS.2026.3708510
\IEEEcompsocthanksitem This work was partially supported by the Research Grants Council of the Hong Kong SAR (Grant No. C5052-23G, PolyU15215623, PolyU15229824, PolyU25216423, PolyU15217424, and SRFS2526-5S04), the National Natural Science Foundation of China (Grant No. 62306259),  and The Hong Kong Polytechnic University (P0056503, P0058445, P0060651).
\IEEEcompsocthanksitem X.~Chen, J.~Wu, C.~Ma, Y.~Yan, and K.~C.~Tan are with the Department of Data Science and Artificial Intelligence, The Hong Kong Polytechnic University, Hong Kong SAR, China.
\IEEEcompsocthanksitem H.~Liu is with the School of Computer Science and Engineering, University of Electronic Science and Technology of China, Chengdu, China.
\IEEEcompsocthanksitem Y.~Wu is with the Department of Computing, The Hong Kong Polytechnic University, Hong Kong SAR, China.
\IEEEcompsocthanksitem Corresponding Author: J.~Wu (jibin.wu@polyu.edu.hk)}}

\maketitle

\begin{abstract} 
Spiking Neural Networks (SNNs) hold great potential to realize brain-inspired, energy-efficient computational systems. However, current SNNs still fall short in terms of multiscale temporal processing compared to their biological counterparts. This limitation has resulted in poor performance in many pattern recognition tasks with information that varies across different timescales. To address this issue, we put forward a novel spiking neuron model called Parallel Multi-compartment Spiking Neuron (PMSN). The PMSN emulates biological neurons by incorporating multiple interacting substructures and allows for flexible adjustment of the substructure counts to effectively represent temporal information across diverse timescales. Additionally, to address the computational burden associated with the increased complexity of the proposed model, we introduce two parallelization techniques that decouple the temporal dependencies of neuronal updates, enabling parallelized training across different time steps. Our experiments across a wide range of pattern recognition tasks demonstrate that PMSN outperforms state-of-the-art spiking neuron models in temporal processing capacity and training speed. Specifically, compared with the commonly used Leaky Integrate-and-Fire neuron, PMSN offers more than 10$\times$ acceleration and a $30\%$ accuracy improvement on Sequential CIFAR-10 dataset, while maintaining comparable computational cost. Our implementation on neuromorphic hardware further demonstrates the deployability of PMSN and highlights its favorable trade-off between effectiveness and efficiency. Therefore, the proposed PMSN presents a promising solution to harness the computational advantages of detailed biological neurons, enabling high-performance and efficient temporal processing on neuromorphic computing systems. Code is available at \url{https://github.com/xychen-comp/PMSN}.
\end{abstract}


\section{Introduction}
\IEEEPARstart{T}{he} human brain, recognized as one of the most sophisticated computational systems on the planet, demonstrates unparalleled energy efficiency and cognitive capabilities. Spiking Neural Networks (SNNs) have been proposed as one of the most representative brain-inspired computational models, aiming to mimic the efficient spatiotemporal information processing in the brain \cite{maass1997networks}. In contrast to traditional artificial neural networks (ANNs) that rely on real-valued neural representations and continuous activation functions, SNNs utilize discrete spike trains to represent information and inherently support efficient event-driven computation. Furthermore, spiking neurons can incorporate rich neuronal dynamics for effective temporal processing \cite{roy2019towards}. At present, SNNs have demonstrated competitive performance and substantial energy savings compared to traditional ANNs \cite{jeffares2022spikeinspired, QCFS, ma2026spatiotemporal} in a wide range of applications, such as image classification \cite{meng2022training, wang2023adaptive}, audio processing \cite{ chen2024hybrid,yan2025efficient}, and robotic control \cite{bdett,chen2025neuromorphic}.

{While SNNs have gained increasing attention in recent years}, their main applications have been primarily confined to computer vision tasks that involve limited temporal dynamics, such as classifying static images or data collected from dynamic vision sensors (DVS) with artificially added saccade motion \cite{li2017cifar10, orchard2015converting}. However, real-world scenarios are often more challenging as they involve sensory signals with information that varies across different timescales. For instance, speech recognition tasks often necessitate models to establish dependencies across various timescales, encompassing phonemes, words, and sentences.
While several algorithms have been developed to enhance the temporal processing capacity of SNNs~\cite{bellec2018long, ALIF, yao2023attention}, most of them still struggle to establish diverse scales of temporal representations, resulting in poor performance in complex temporal processing tasks.

Considering the remarkable temporal processing capabilities observed in biological neurons, it is crucial to carefully examine their computational mechanisms. In the context of large-scale SNNs, the majority of existing spiking neurons are modeled as single-compartment systems, such as the Leaky Integrate-and-Fire (LIF) model \cite{lif}. In these single-compartment models, the neuronal dynamics are simplified to first-order dynamics of a single state variable - the membrane potential, disregarding any interactions among substructures within a single neuron. This simplification reduces computational complexity but limits their ability to generate complex neuronal dynamics.

In contrast, real biological neurons can be better modeled by multi-compartment neuron models \cite{hines1984efficient, spruston2008pyramidal, gidon2020dendritic}, {which divide a single neuron into several interconnected subunits with interactions among them}. Extensive neuroscience experiments have demonstrated the important roles of these {interactive} dynamics in temporal processing. For example, the nonlinear interactions among ion channels endow neurons with significant computational power for processing temporal signals \cite{poirazi2003pyramidal, major2013active}. 
Additionally, the interaction among coupled dendritic components acts as temporal filters for signals traveling from dendrites to the soma, thereby facilitating the temporal sequence detection \cite{rall1964theoretical, spruston2008pyramidal}. 
While detailed multi-compartment models show great potential in temporal processing, they are computationally expensive due to the intricate anatomical structures of dendritic trees and high-dimensional ionic properties. Consequently, they are not well suited for constructing large-scale SNNs to tackle real-world pattern recognition tasks. Therefore, there is a pressing need to develop an efficient spiking neuron model that strikes a balance between computational complexity and the ability to capture the valuable interaction among different substructures of a biological neuron.

Recently, several simplified compartmental spiking neuron models have been specifically designed for deep SNNs. For instance, a two-compartment model was introduced to simulate double-exponential threshold decay \cite{DEXAT}, and a dendritic neuron model was proposed to endow multiple dendritic compartments with heterogeneous decaying time constants \cite{zheng2024temporal}. Furthermore, inspired by the well-known Pinsky-Rinzel (P-R) pyramidal neuron located in the CA3 region of the hippocampus, a two-compartment LIF model was proposed, which divides a single neuron into dendritic and somatic compartments~\cite{zhang2023tclif}. 

Nevertheless, there are three significant challenges that remain to be tackled. Firstly, these models overlook the valuable recurrent interactions among neuronal compartments, {which play a crucial role in integrating temporal information across multiple timescales in biological neurons \cite{poirazi2020illuminating,acharya2022dendritic}. As a result, the expressibility of these models is limited, constraining the complexity of temporal dynamics they can effectively capture.} Secondly, {these meticulously handcrafted models are inherently constrained by a predetermined number of compartments. While increasing this number allows for richer structural complexity and a broader range of temporal dynamics, reducing it yields a more compact and computationally efficient representation. However, the inability to adjust this abstraction level limits these models' adaptability to tasks with varying temporal complexity and computational demands.} Thirdly, compared to single-compartment models, the increased temporal dynamics in these models result in significantly slower training processes when using the back-propagation through time (BPTT) algorithm. Consequently, these models demonstrate limited scalability to larger networks for tackling challenging real-world temporal processing tasks, particularly those involving long sequences.

To tackle these challenges, we propose a generalized multi-compartment spiking neuron model for SNNs that integrates essential recurrent interactions among interconnected compartments. Moreover, the number of compartments can be flexibly adjusted in the proposed model, allowing for adaptation to the different temporal complexities required in real-world tasks.
Furthermore, considering the significant constraint on training speed caused by increasingly complex neuronal dynamics in the proposed model, we introduce two temporal parallelization techniques to accelerate the training process. 
{Firstly, for the linear recurrence in the neuron charging process, we model it as a linear time-invariant (LTI) system, which can be efficiently parallelized over time following existing approaches for linear recurrent models \cite{martin2018parallelizing,s4,s5}.
Furthermore, for the nonlinear recurrence associated with the neuron firing and reset process, we introduce a novel reset strategy. This mechanism effectively decouples the nonlinear temporal dependency, enabling parallel computation that is mathematically equivalent to a step-by-step, spike-triggered reset mechanism.}
{Altogether, these strategies enable efficient temporal parallelization during training, while preserving the event-driven, online inference on neuromorphic hardware.}

\begin{figure*}
    \centering \includegraphics[width=1\linewidth]{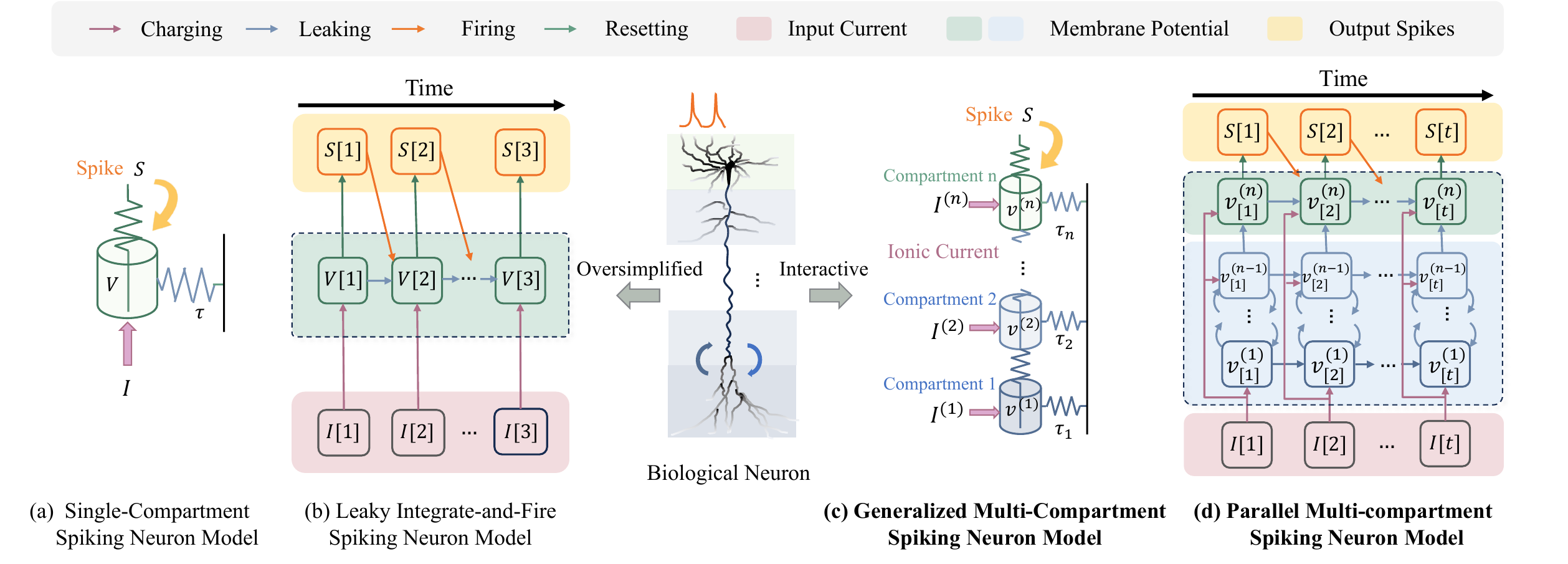} 
    \caption{{Comparison of neuronal structure and dynamics between the single-compartment model, biological neurons, and the proposed multi-compartment spiking neuron model. \textbf{(a, b)} The widely used LIF model simplifies the biological neurons into a single unit and ignores the interaction among neuronal substructures, resulting in deficiencies in multiscale temporal processing and slow training speed. \textbf{(c)} In contrast, the detailed morphologies and electrophysical properties of biological neurons improve their computational power in interactive ways, which is crucial for temporal information processing. Drawing inspiration from these, we propose a generalized multi-compartment spiking neuron model that divides a single neuron into a variable number of interconnected subunits with heterogeneous temporal properties and endows them with interactive dynamics. \textbf{(d)} Our proposed PMSN model further extends the compartmental structure in (c). It not only {supports rich} neuronal dynamics essential for temporal processing, but also facilitates efficient parallel training across time.
    }}
    \label{fig: main}
\end{figure*}

The proposed design methodologies have led to the development of a novel Parallel Multi-compartment Spiking Neuron (PMSN) model as illustrated in Fig. \ref{fig: main}. The PMSN model effectively captures multiscale temporal information through the interaction among neuronal compartments. Both theoretical analysis and dynamics visualizations are provided to substantiate the efficacy of the PMSN model in establishing temporal dependencies across various timescales. Moreover, the proposed PMSN model offers a significant improvement in training speed, particularly when deployed on GPU-accelerated machine learning (ML) frameworks. The main contributions of this work are summarized as follows:

\begin{itemize}[leftmargin=*]
    \item {We propose a generalized multi-compartment spiking neuron model for deep SNNs that incorporates valuable interactions among different neuronal compartments. The number of compartments in this model can be flexibly adjusted to represent temporal information across diverse timescales.}
    \item We develop two temporal parallelization techniques for the proposed model, achieving more than a $10\times$ training speedup on GPU-accelerated ML frameworks.
    \item {Our comprehensive experiments demonstrate the superior temporal processing capacity and substantial training acceleration of PMSN, as well as its potential for energy-efficient deployment on neuromorphic platforms.}
    
\end{itemize}

\section{Related Works}
\label{sec:related}
\subsection{Advances in Spiking Neuron Models}
Recent enhancements in spiking neuron models for temporal processing can be categorized into single- or {multi}-compartment models according to the number of membrane potential subunits involved \cite{markram2006blue}. Several
single-compartment models have incorporated adaptive variables to enhance the efficacy of temporal information representation.
For instance, \cite{bellec2018long} and \cite{ALIF} propose to use adaptive firing thresholds, which function as long-term memory to enhance temporal processing; 
\cite{PLIF} utilizes learnable decay constants to enhance the heterogeneity of neuron populations, enabling them to effectively represent multiscale temporal information; 
\cite{GLIF} introduces a gating mechanism into the neuron model to explicitly control memory storage and retrieval processes. 

Comparatively, {multi-compartment models utilize compartmental heterogeneity or recurrent} interaction to represent information across different timescales.
For example, \cite{DEXAT} proposes a two-compartment model where the thresholds undergo a double-exponential decay, facilitating the storage of both short- and long-term information; \cite{zheng2024temporal} introduces a heterogeneous model consisting of one somatic compartment and varied dendrite-branch compartments; \cite{zhang2023tclif} proposes a two-compartment model that captures the interactive dynamics between the soma and dendrites. Although these handcrafted models demonstrate enhanced performance, there is a pressing need for a principled approach to efficiently scale them and incorporate a variable number of compartments. This is crucial for enhancing their temporal processing capability, which serves as the central focus of this paper.

\subsection{{Training Techniques for SNNs}}

{Training SNNs poses fundamental challenges for gradient-based optimization due to the non-differentiable nature of spike generation \cite{eshraghian2023training}. Existing training methods for deep SNNs can be broadly categorized into conversion-based and direct training approaches. Conversion-based methods transfer knowledge from pre-trained ANNs to SNNs by matching activation or firing-rate statistics \cite{QCFS,LTL}. While effective for static tasks, such approaches are generally unsuitable for temporal processing, as they introduce non-negligible approximation errors in temporal dynamics. In contrast, direct training methods optimize SNNs end-to-end by approximating spike gradients with surrogate functions, enabling effective learning on temporal signals \cite{li2021differentiable,yang2023effective}. {
Recent advances have further demonstrated their potential in efficient language modeling~\cite{snnbert}, robust visual recognition~\cite{snnrat}, and event-based optical flow estimation~\cite{yang2024self}.} However, due to the recurrent nature of spiking neuron dynamics, most direct training approaches rely on BPTT, or its online variants \cite{xiao2022online}, resulting in training complexity that scales linearly with sequence length. This severely limits their scalability to long sequences.}

{Several recent works have explored parallelization strategies to alleviate this limitation. Differentiation on Spike Representation (DSR) reformulates spiking dynamics using differentiable rate-based representations, enabling constant-time gradient computation \cite{meng2022training}. However, this approach is restricted to simple Integrate-and-Fire (IF) or LIF neurons and relies on rate-based representations, making it unsuitable for modeling fine-grained temporal dependencies. In parallel, the Parallel Spiking Neuron (PSN) family \cite{fang2023parallel} reformulates membrane potential charging dynamics using a learnable decay matrix and omits the reset mechanism to enable temporal parallelism. Despite their efficiency, PSN models require access to future inputs and are limited to single-compartment structures, making them biologically implausible, incompatible with neuromorphic hardware, and difficult to extend to richer multi-compartment temporal modeling.}

\section{Revisit: Single-compartment Spiking Neuron Models Struggle to Process Temporal Signals}
\label{preliminary}
In this section, we revisit the LIF model as a representative of single-compartment spiking neuron models. We first introduce basic concepts and then discuss the major challenges associated with processing temporal signals using this model. As illustrated in Fig. \ref{fig: main}, the temporal dynamics of the LIF model can be formulated as:
\begin{equation}
\begin{aligned}
&\text{Leak \& Charge:} \quad \frac{\mathrm{d} v(t)}{\mathrm{d}t}=-\frac{1}{\tau_m }(v(t)-v_{rest})+I(t),\\
&\text{Fire \& Reset:} \quad \text{if} \ v(t)\geq \theta, s(t)=1, {\ v(t)\rightarrow v(t)-\theta}.
\end{aligned}
\end{equation}

During the membrane potential leaky and charging phase, the information contained in the input current $I(t)$ is integrated into the membrane potential $v$ and further undergoes decay at a rate governed by $\tau_m$. $v_{rest}$ is the resting potential. Once $v(t)$ exceeds the firing threshold $\theta$, an output spike will be generated and transmitted to subsequent neurons, after which the membrane potential is reset. In practice, the above continuous-time formulation is typically discretized using the Euler method as
\begin{equation}
\label{eq:lif}
\left\{\begin{array}{lr}
V[t]=\alpha V[t-1] +I[t] -\theta S[t-1],& \\
S[t]=H\left(V[t] {-}\theta\right), &
\end{array}\right.
\end{equation}
where $H(\cdot)$ is the Heaviside function and $\alpha=\mathrm{exp}(-\frac{\mathrm{d}t}{\tau_m})$ is the discrete-time decay factor. {Following this discretization, LIF neuron dynamics are simulated via recurrent state updates over time, where the membrane potential $V[t]$ depends on its previous state $V[t{-}1]$ as well as the current input. More generally, such recurrent evolution is a fundamental characteristic of SNNs, allowing them to be naturally interpreted as recurrent models with intrinsic temporal dependencies.}

Despite its promising results in tasks involving limited temporal context, LIF neurons encounter the following two challenges when dealing with long sequences. Firstly, this model faces challenges in retaining information over an extended time period, primarily due to its inadequate representation of neuronal dynamics. Specifically, the LIF model overlooks the computational role played by interactions among different neuronal substructures. As a result, the membrane potential $V$ is the only state variable that can integrate and store temporal information. Unfortunately, this state variable is subject to exponential decay and reset, which hinders the establishment of long-term temporal dependencies. 
{Secondly, its training time increases proportionally with the sequence length, making it inefficient for tasks that involve long sequences. This slow training speed arises from the nonlinear recurrent updates in the neuronal dynamics, which must be computed step by step along the temporal dimension as commonly observed in RNNs, leading to underutilization of the full potential of GPU acceleration. Specifically, the membrane potential update of $V[t]$ depends on the output spike $S[t-1]$ from the preceding timestep, which is generated by applying a nonlinear Heaviside activation to $V[t-1]$. As a result, $V[t]$ is not available until $V[t-1]$ has been fully processed. This time-coupled relationship prevents the membrane potential dynamics from being unfolded across time, leading to difficulties in temporal parallelization.}

\section{A Generalized Multi-compartment Spiking Neuron Model with Interactive Dynamics}
{In this section, we first briefly introduce the multi-compartment models that have been widely used in neuroscience studies. Drawing inspiration from them, we propose a generalized multi-compartment spiking neuron model that can strike a favorable balance between the richness of neuronal dynamics and computational cost. Notably, in contrast to commonly used multi-compartment neuron models in deep SNNs, our model retains the crucial recurrent interactions among interconnected compartments. It thereby allows more effective representation and processing of temporal signals across different timescales.}

{Compartmental models divide a single neuron into interconnected subunits or ``compartments'', with specific spatial structures. 
The interactions among these interconnected compartments lead to the rich neuronal dynamics observed in biological neurons that play a crucial role in sensory processing \cite{buzsaki2012mechanisms}. 
For instance, Rall's cable theory \cite{rall1964theoretical} suggests each dendrite can be mathematically modeled as a series of interconnected compartments resembling a cable structure.}
However, given the complex anatomical morphology of dendritic trees, these models pose significant computational challenges and are impractical for large-scale simulations. To reduce the modeling difficulty, quantitative models with reduced compartment numbers and structure complexity have been explored. For instance, the neuronal activities of CA3 pyramidal neurons can be modeled by a 19-compartment cable model, wherein dendritic branching is omitted \cite{traub1991model}. This model has been further abstracted to a more computationally efficient two-compartment model \cite{pinsky1994intrinsic}. In these quantitative models, compartments are restricted to interacting with their adjacent compartments solely, and the dynamics of each compartment can be described as a differential equation of a resistor-capacitance circuit as
\begin{equation}
\label{eq: real_neuron}
\begin{aligned}
    C_m \frac{\mathrm{d} v^{(i)}}{\mathrm{d} t}=& -I_{Leak}( v^{(i)}) + g_{i-1,i}(v^{(i-1)}-v^{(i)})\\
    &+ g_{i,i+1}(v^{(i+1)}-v^{(i)}) + I.
\end{aligned}
\end{equation}
{where $C_m$ is the capacitance of membrane potential, $I_{Leak}(\cdot)$ represents the leakage current of membrane potential $v^{(i)}$, $g_{i,j}(\cdot)$ is the coupling conductance between compartments $i$ and $j$, and $I$ denotes the summation of other voltage-gated ionic currents.}

{In order to simplify the ionic compartmental dynamics described in Eq. \eqref{eq: real_neuron} and ensure compatibility with existing GPU-accelerated ML frameworks, we next introduce a generalized multi-compartment spiking neuron model with a simplified neuronal hyperparameter setup.}
This model retains the essential interactions among different substructures of a neuron, {which are theoretically evidenced in Section \ref{gradient_update} as playing a crucial role in capturing and integrating temporal features across multiple timescales.}
Another notable feature of our model is its flexibility, allowing for a varying number of neuronal compartments $n$ to suit the complexity of the task at hand. 
The detailed model formulation is given as 
\begin{equation}
\begin{aligned}
\label{eq:generalized}
&\left\{\begin{array}{lr}
\frac{\mathrm{d} v^{(1)}(t)}{\mathrm{d} t}=-\frac{1}{\tau_1}v^{(1)}(t)  +\beta_{2,1}v^{(2)}(t) + \gamma_1 I(t), & \\
\frac{\mathrm{d} v^{(2)}(t)}{\mathrm{d} t}=-\frac{1}{\tau_2}v^{(2)}(t)+\beta_{3,2}v^{(3)}(t)+\beta_{1,2}v^{(1)}(t)+\gamma_2 I(t), & \\
 \qquad \qquad \qquad ... &\\
\frac{\mathrm{d} v^{(n)}(t)}{\mathrm{d} t}=-\frac{1}{\tau_n}v^{(n)}(t)+\beta_{n-1,n}v^{(n-1)}(t)+\gamma_n I(t), & 
\end{array}\right.\\
&\begin{array}{ll}
\text{if} \ &v^{(n)}(t)\geq \theta, \quad \ s(t)=1, \quad v^{(n)}(t)\leftarrow v^{(n)}(t)-\theta,
\end{array}
\end{aligned}
\end{equation}
where $v^{(i)}$ represents the membrane potential of the compartment $i$, $\theta$ is the firing threshold. $I^l(t)=\mathcal{W}^l S^{l-1}(t)$ denotes the synaptic current induced by the output spikes of the preceding layer, where $\mathcal{W}^l$ represents the synaptic weight between layer $l-1$ and $l$. Once the membrane potential of the final compartment $v^{(n)}$ exceeds the threshold $\theta$, it triggers an output spike and simultaneously resets $v^{(n)}$. The compartmental parameters, including $\tau_i$, $\gamma_i$, and $\beta_{i,j}$, represent the membrane time constant, input attenuation of the compartment $i$, and coupling strength between interconnected compartments $i$ and $j$, respectively.

{For the sake of simplicity in representation, these first-order differential equations can be equivalently expressed as an $n$-dimensional first-order state-space dynamical system}: 
\begin{equation}
\label{eq:multicompartment}
\begin{aligned}
\dot{\mathcal{V}}(t)&= \begin{bmatrix}
-\frac{1}{\tau_1} & \!\beta_{2,1}\! & \!0\! & \!\mathclap{\cdots}\! & \!0\! \\
\!\beta_{1,2}\! & \!-\frac{1}{\tau_2}\! & \!\beta_{3,2}\! & \!\mathclap{\cdots}\! & \!0\! \\
\!\mathclap{\vdots}\! & \!\mathclap{\vdots}\! & \!\mathclap{\vdots}\! & \!\ddots\! & \!\mathclap{\vdots}\! \\
\!0\! & \!0\! & \!\mathclap{\cdots}\! & \!-\frac{1}{\tau_{n-1}}\! & \!\beta_{n,n-1}\! \\
\!0\! & \!0\! & \!\mathclap{\cdots}\! & \!\beta_{n-1,n}\! & \!-\frac{1}{\tau_n}\! \\
\end{bmatrix}\mathcal{V}(t) + \boldsymbol{\gamma_n} I(t),\\
 S(t)&= H(v^{(n)}(t) - \theta ), \qquad v^{(n)}(t) \leftarrow  v^{(n)}(t) - \theta S(t), 
\end{aligned}
\end{equation}
where $\boldsymbol{\gamma_n}=[\gamma_1, \cdots, \ \gamma_n]^T$, $\mathcal{V}=[v^{(1)} , \cdots, \ v^{(n)}]^T$.

{It is important to note that many widely used spiking neuron models can be seen as special cases of our proposed model by adjusting its hyperparameters. For instance, when the number of compartments $n$ is set to $1$ and $\beta=0$, our model is simplified to a standard LIF model. Similarly, when $n=2$ and $\gamma_2=0$, our model reduces to a Two Compartment LIF (TC-LIF) model \cite{zhang2023tclif}. These examples highlight the generalizability of our model. Moreover, by providing flexibility in increasing the number of compartments, our model surpasses existing spiking neuron models and offers richer neuronal dynamics that are essential for complex temporal processing tasks.}

\begin{figure*}[t]
\centering
\includegraphics[width=0.9\linewidth,trim= 0 10 0 0, clip]{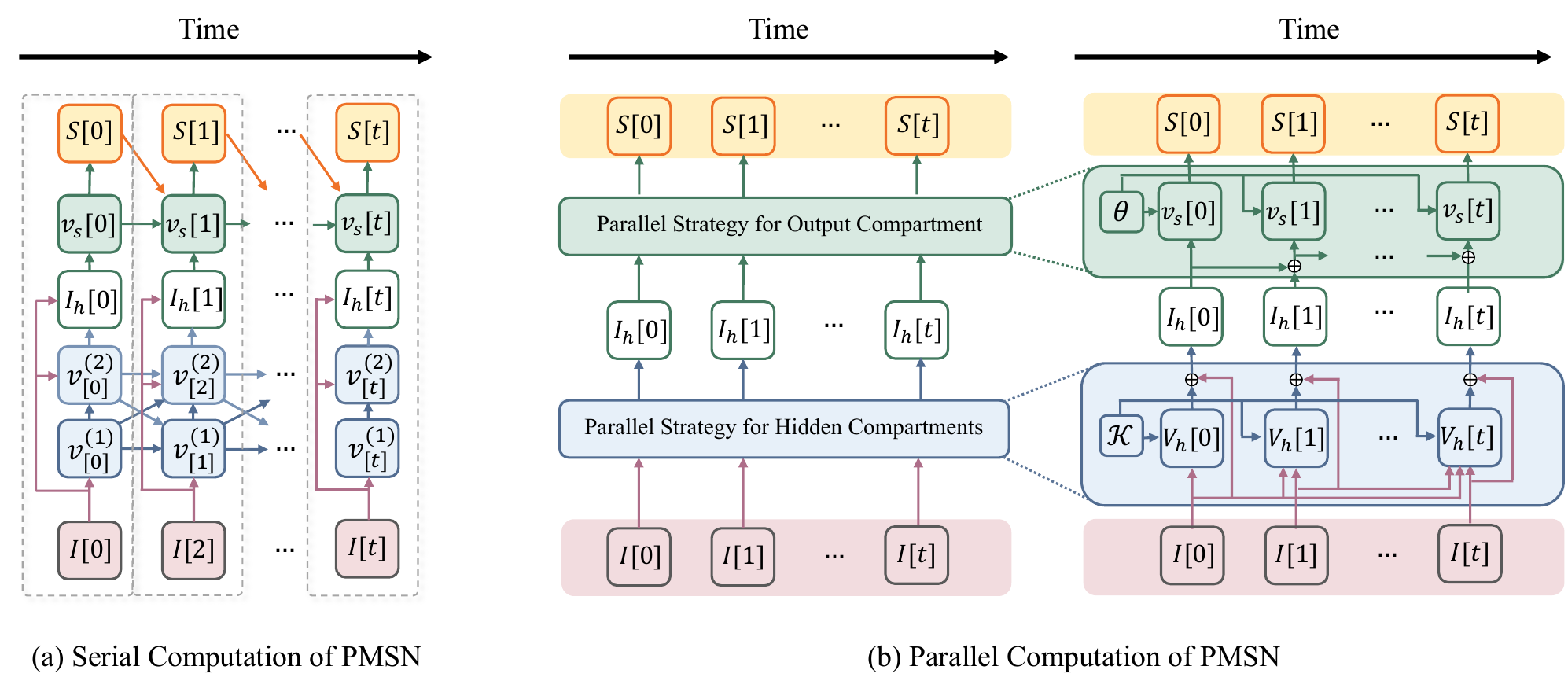}\hspace{3em}
\caption{{Illustration of the proposed PMSN model and its parallel implementation. \textbf{(a)} The PMSN model ($n=3$ compartments for demonstration) can be divided into two parts: two hidden compartments with membrane potential $\mathcal{V}_h=[v^{(1)}, ~v^{(2)}]$, and one output compartment with membrane potential $v_s$. $I,~I_h,~S$ represent the input current, total input to the output compartment, and output spikes, respectively. Due to the nonlinear temporal dependencies among these states, calculating neuronal temporal dynamics typically involves serial computation over time, resulting in significant training costs that scale proportionally with sequence length. \textbf{(b)} The proposed parallel solution. \textbf{Left:} To accelerate the training speed, two temporal parallel strategies are introduced for the hidden (blue box) and output compartments (green box), respectively. \textbf{Right:} The schematic of each parallel block. For the hidden compartments, we decouple the linear recurrence between $v^{(1)}, ~v^{(2)}$, enabling membrane potentials $V_h$ to be computed in parallel across time. This is achieved by applying a temporal convolution between the input current $\boldsymbol{I_t}$ and a kernel $\mathcal{K}$ derived from the underlying neuronal dynamics.
For the output compartment, we propose a novel reset mechanism TDR that leverages the firing threshold $\theta$ and the accumulated inputs $I_h$ to compute output spikes at all timesteps simultaneously.}}
\label{fig: parallel}
\end{figure*}

\section{PMSN: A Parallel Multi-compartment Spiking Neuron}

The generalized multi-compartment spiking neuron model introduced earlier requires significantly more training time compared to existing spiking neuron models used in deep SNNs, primarily due to its higher computational complexity. This limitation restricts its practical deployment in long sequence temporal processing scenarios. Thus, it becomes crucial to develop strategies to enable parallel training of this model in time. However, it is not straightforward to employ existing parallel computation techniques \cite{martin2018parallelizing}, due to the involvement of the nonlinear function $H(\cdot)$ for spike generation and reset processes. 

In this section, we further propose a {PMSN} model
with two parallelization techniques independently designed for hidden and output compartments. To better explain our idea, we divide the total $n$ compartments into $n-1$ hidden compartments $\mathcal{V}_h$ with linear recurrence and one output compartment $v^{(n)}=v_s$ with nonlinear firing and reset dynamics. Meanwhile, the feedback from the output to the last hidden compartment $\beta_{n,n-1}$ is set to $0$. The resulting neuronal function can be represented as

\begin{equation}
\label{eq:enhance_h}
\dot{\mathcal{V}}_h(t)= \begin{bmatrix}
-\frac{1}{\tau_1} & \!\beta_{2,1}\! & \!0\! & \!\mathclap{\cdots}\! & \!0\! \\
\!\beta_{1,2}\! & \!-\frac{1}{\tau_2}\! & \!\beta_{3,2}\! & \!\mathclap{\cdots}\! & \!0\! \\
\!\mathclap{\vdots}\! & \!\mathclap{\vdots}\! & \!\mathclap{\vdots}\! & \!\ddots\! & \!\mathclap{\vdots}\! \\
\!0\! & \!0\!  & \!0\! & \!\mathclap{\cdots}\!  & \!-\frac{1}{\tau_{n-1}}\! \\
\end{bmatrix} \mathcal{V}_h(t) + \boldsymbol{\gamma_{n-1}} I(t),
\end{equation}
\begin{equation}
\label{eq:enhance_s}
     \dot{v}_s(t)=\beta_{n-1,n} v^{(n-1)}(t) -\frac{1}{\tau_{n}}v_s(t)+\gamma_n I(t)-\theta S(t),
\end{equation}
\begin{equation}
\label{eq:enhance_re}
     S(t)= H(v_s(t) - \theta ).
\end{equation}

Below, we will introduce how these two parallelization techniques unfold the linear recurrent states $\mathcal{V}_h$, and decouple the nonlinear temporal dependency issue associated with $v_s$ to enable parallel computation.

\subsection{Parallel Strategy for Hidden Compartments}

{By examining the hidden compartment dynamics of PMSN in Eq.~\eqref{eq:enhance_h}, we observe that the hidden states $\mathcal{V}_h$ evolve according to an LTI dynamical system. Specifically, the membrane potential of each hidden compartment $v^{(i)}$ is updated by applying a time-invariant state transition matrix to its previous state. As a result, the hidden compartment dynamics admit a state-space formulation, which shares a common underlying mathematical structure with linear recurrent neural networks (RNNs) \cite{martin2018parallelizing, orvieto2023resurrecting} and state-space models (SSMs) \cite{s4,s5} widely used in modern sequential modeling.
This connection bridges biologically grounded neuron dynamics with contemporary deep learning architectures and, importantly, allows PMSN to leverage parallel computation techniques originally developed for parameterized SSMs to efficiently compute the temporal evolution of its hidden compartments.}

We first apply the zero-order hold (ZOH) method \cite{decarlo1989linear} to discretize the nonlinear continuous system detailed in Eqs. \eqref{eq:enhance_h} and \eqref{eq:enhance_s}. 
Specifically, we utilize a full-rank state transition matrix $\mathcal{T}\in \mathbb{R}^{ (n-1)\times (n-1) }$ to represent the first matrix in Eq. \eqref{eq:enhance_h}, and diagonalize {this matrix} via eigenvalue decomposition $\mathcal{T}=P\Lambda P^{-1} $, where $\Lambda, P \in  \mathbb{C}^{(n-1)\times (n-1)}$ denote the diagonal eigenvalue matrix and eigenvector matrix, respectively. We then obtain the following discrete-time formulation:
\begin{equation}
\label{eq:model_woreset1}
{V}_h[t]= \Bar{\mathcal{T}} {V}_h[t-1] + {{\Phi}_c} I[t], \qquad \qquad 
\end{equation}
\begin{equation}
\label{eq:I_h}
    I_h[t]={\Phi}_s {V}_h[t] + \gamma_n I[t], \qquad \qquad \quad \
\end{equation}
\begin{equation}
\label{eq:model_woreset2}
\begin{aligned}
    &{v_s}[t]=\alpha v_s[t-1] + I_h[t] -\theta S[t-1], \\  &S[t]= H(v_s[t] - \theta ),
\end{aligned}
\end{equation}
where $V_h=P^{-1}\mathcal{V}_h$, $\Bar{\mathcal{T}}=\mathrm{exp}(\Lambda \mathrm{d}t)$, ${\Phi}_c=\Lambda^{-1}(\mathrm{exp}(\Lambda \mathrm{d}t)-I)\phi_c$, and $\phi_c=P^{-1}\boldsymbol{\gamma_{n-1}}.$ The term $I_h[t]$ signifies the total input to the output compartment, ${\Phi}_s=[0,..,\beta_{n-1,n}]P$, $\alpha = \mathrm{exp}(-\frac{\mathrm{d}t}{\tau _n})$. 
The parameters $\Lambda \mathrm{d}t,~\phi_c,~\gamma_n$, and $\Phi_s$ are learnable. It should be noted that these complex number operations are well-supported by existing neuromorphic chips, such as Intel Loihi \cite{davies2021taking}.

The model described in Eq. \eqref{eq:model_woreset1} exhibits a linear recurrence that can be unfolded over time as
\begin{equation}
    {V}_h[t]=\sum^t_{i=0}{\Bar{\mathcal{T}}^{t-i}{{\Phi}_c}I[i] }.
\label{eq:v_h_t}
\end{equation}

By substituting Eq. \eqref{eq:v_h_t} into Eq. \eqref{eq:I_h}, we can obtain
\begin{equation}
\label{eq:I_h_2}
I_h[t]=\sum^t_{i=0}\Phi_s\Bar{\mathcal{T}}^{t-i}{{\Phi}_c}I[i]  +\gamma_n I[t].
\end{equation}

{Notably, the first term in Eq. \eqref{eq:I_h_2} can be simplified to a convolution form as}
\begin{equation}
\label{eq:I_h_3}
    I_h[t] = \sum_{i=0}^t\boldsymbol{I_t}[i]\mathcal{K}[t-i] +\gamma_n I[t] =(\boldsymbol{I_t} * \mathcal{K})[t]  +\gamma_n I[t],
\end{equation}
{where $\boldsymbol{I_t}=\{I[0],..,I[t]\}$ denotes the input current set, $\mathcal{K}=[\Phi_s\Bar{\mathcal{T}}^{0}\Phi_c,\ ...,\ \Phi_s\Bar{\mathcal{T}}^{t}\Phi_c]$ is the convolution kernel.  
Consequently, the computation of set $\boldsymbol{I_h}=\{I_h[0],...,I_h[t]\}$ can be parallelized over time by applying a convolution operation on $\boldsymbol{I_t}$ and $\mathcal{K}$, resulting in}
\begin{equation}
\label{eq:fft}
    \boldsymbol{I_h} = \boldsymbol{I_t} * \mathcal{K}   + \gamma_n 
 \boldsymbol{I_t}= \mathcal{F}^{-1}(\mathcal{F}(\boldsymbol{I_t})\cdot \mathcal{F}(\mathcal{K}))+\gamma_n \boldsymbol{I_t},
\end{equation}
where $\mathcal{F},~\mathcal{F}^{-1}$ represent forward and inverse Fourier transforms, respectively. 
In this way, we could efficiently compute the membrane potential for the first $n-1$ hidden compartments $V_h$, and the input for the output compartment $I_h$ across all time steps in parallel. 

\subsection{Parallel Strategy for Output Compartment with Reset}
{In the previous subsection, we demonstrated that the dynamics of PMSN within the hidden compartments can be efficiently computed in parallel. The remaining question is how to achieve parallel computation in the output compartment. }

{However, achieving temporal parallelization for the output compartment is non-trivial due to the spike generation and reset mechanism, which introduces a nonlinear dependency across consecutive timesteps. Specifically, in conventional reset formulations in Eq.~\eqref{eq:model_woreset2}, the membrane potential $v_s[t]$ is updated based on the previous spike $S[t-1]$, where $S[t-1]$ is generated by applying a non-differentiable Heaviside function to $v_s[t-1]$. This spike-triggered reset tightly couples $v_s$ across time, preventing the membrane dynamics from being unfolded into a closed-form expression and fundamentally limiting temporal parallelization.}

{To address this limitation, we propose a novel reset mechanism termed temporal-decoupled reset (TDR). TDR reformulates the reset operation such that it can be evaluated either in a serial (step-by-step) manner or in a temporally parallel manner, while preserving the same spike-driven reset behavior.
Concretely, we introduce a reset term $v_r[t]$ that represents the discharged membrane voltage after spiking. In the serial form, $v_r[t]$ is expressed as an integer multiple of the firing threshold $\theta$, determined by the membrane potential $v_s[t]$. Its dynamics can be formulated as}

\begin{equation}
\label{eq:modified_out}
\begin{aligned}
    &v_s[t]= {v}_s[t-1]+I_h[t]  - v_r[t-1], \\ &S[t]=H(v_s[t] - \theta),\\
&v_r[t]= S[t] \cdot \lfloor{ {v}_s[t]}\rfloor_{\theta},
\end{aligned}
\end{equation}
where $ \lfloor x \rfloor_{\theta}:=\theta \lfloor x/\theta \rfloor$ signifies the floor division by $\theta$. 

\xy{Similar to the commonly used reset mechanisms in conventional spiking neuron models, TDR can be viewed as a biologically inspired reset mechanism that abstracts the depolarization--fire--repolarization process into a simplified form suitable for efficient computation and neuromorphic hardware implementation. Specifically, soft reset removes one fixed threshold after spike generation, while hard reset clamps the membrane potential to a reset value. Following the same abstraction, TDR differs only in the amount of membrane voltage discharged after spike generation: it removes a threshold-quantized voltage determined by the floor operation, which helps keep the post-reset membrane potential within the subthreshold range.}
{Importantly, the same reset dynamics can also be computed in a temporally parallel form. Specifically, the accumulated discharged voltage $\sum_{i=0}^{t-1} v_r[i]$ can be directly derived from the cumulative input current $\sum_{i=0}^{t-1} I_h[i]$ across time, thereby bypassing the evolution of the membrane potential $v_s[t]$ as}

\begin{equation}
\begin{aligned}
\label{eq:sumreset}
   &\sum^{t-1}_{i=0} v_r[i]=  \lfloor{\sum^{t-1}_{i=0} I_h[i]}\rfloor_{\theta}, \\ 
   &v_s[t]= \sum^{t}_{i=0} I_h[i] - \lfloor{\sum^{t-1}_{i=0} I_h[i]}\rfloor_{\theta},
\end{aligned}
\end{equation} 

A detailed derivation is provided in the \textit{Supplementary Materials}. To ensure equivalence between Eq. \eqref{eq:sumreset} and Eq. \eqref{eq:modified_out}, the input current $I_h$ is passed through a rectified linear unit (ReLU) to enforce non-negativity before being injected into the output compartment. {As a result, the parallel formulation of TDR can be expressed by directly deriving the output spike train $\boldsymbol{S_t}=\{S[0],..,S[t]\}$ from the corresponding membrane potential sequence $\boldsymbol{v_{s,t}}=\{v_s[0], ..., v_s[t]\}$ as}
\begin{equation}
    \boldsymbol{S_t} = H\left( \boldsymbol{v_{s,t}}\right)=H\left(\boldsymbol{I_{h,t}}- \lfloor{\boldsymbol{I_{h,t-1}}}\rfloor_{\theta} \right).
\label{eq: output_parallel}
\end{equation}
{where the set $\boldsymbol{I_{h,t}}=\{ I_h[0],...,\sum^{t}_{i=0} I_h[i]\}$, representing the cumulative input current to the output compartment over time, can be efficiently computed using the parallel prefix sum (Scan) algorithm \cite{harris2007parallel}. }

\begin{table*}[t]
  \caption{{Comparison of the proposed reset mechanism in PMSN with existing reset mechanisms used in other spiking neuron models.}}
  \label{tab: reset_compare}
  \centering
  \renewcommand{\arraystretch}{1.25}
  \resizebox{0.98\textwidth}{!}{%
  {
  \begin{tabular}{lccccc}
    \toprule
    \textbf{Reset Mechanism} & \textbf{Typical Models} & \textbf{Formulation} & \makecell[c]{\textbf{Parallelizable}\\ \textbf{Training}} & \makecell[c]{\textbf{Equivalent} \\ \textbf{Online Inference}} & \makecell[c]{\textbf{Spike} \\ \textbf{Sparsity}} \\
    \midrule
     {Soft Reset (Reset by subtraction)} & LIF \cite{diehl2016truehappiness}, ALIF \cite{ALIF} & $v[t]  \leftarrow v[t] -\theta$ & \ding{55} & - & ++\\
    
     {Hard Reset (Reset to zero)} & LIF \cite{diehl2015fast}, PLIF \cite{PLIF} & $v[t] \leftarrow 0$ & \ding{55} & - & +++ \\

    {Fused Reset} & GLIF \cite{GLIF} & $v[t] \leftarrow (1-\gamma)(v[t]-\theta)$ & \ding{55} & - & ++ \\

    {Reset-free} & PSN \cite{fang2023parallel}, P-SpikeSSM \cite{bal2025pspikessm}  &  $v[t]\leftarrow v[t]$ & \ding{51}  & \ding{51}  & +\\
    
    {Surrogate Dynamic Network (SDN)} & SpikingSSMs \cite{shen2025spikingssms}  &  $v[t]\leftarrow f(I[0:t],\theta)$ & \ding{51}  & \ding{55} & +++ \\
    \midrule
      \textbf{TDR (Ours)} & \textbf{PMSN}  & $v[t]\leftarrow v[t]- \lfloor v[t]\rfloor_\theta$ & \ding{51} & \ding{51} & +++\\
    \bottomrule
     \multicolumn{6}{l}{$v$ - membrane potential, $\theta$ - threshold, $I$ - input, $\gamma \in (0,1)$ - gating factor,  $f(\cdot)$ - learnable MLP-based function predicting reset values from inputs.}
  \end{tabular}}}
\end{table*}

Compared with existing reset mechanisms summarized in Table~\ref{tab: reset_compare}, the proposed TDR exhibits several distinctive advantages. Unlike soft reset, hard reset, or fused reset strategies, which inherently rely on spike-dependent, step-by-step updates, TDR enables temporally parallel computation during training, substantially improving training efficiency. In contrast to reset-free or Surrogate Dynamic Network (SDN) approaches \cite{shen2025spikingssms}, TDR preserves spike sparsity and maintains equivalent online, event-driven inference behavior, ensuring full compatibility with neuromorphic hardware deployment. \xy{Moreover, the proposed reset mechanism can be generalized to other single-compartment neuron models or multi-compartment models with multiple spike-generating output compartments, thereby enabling parallel training for a broader class of spiking neuron architectures.}

\begin{table*}[t]
\caption{\xy{Comparison of classification accuracies on neuromorphic sequential tasks.}} 
\label{tab: mnist}
\renewcommand{\arraystretch}{1.0}
\begin{center}
\resizebox{0.9\textwidth}{!}{
\begin{tabular}{ccccccc}
\toprule
\multicolumn{1}{c}{\bf Dataset}  & \multicolumn{1}{c}{\bf Seq. Length} & \multicolumn{1}{c}{\bf Approach} &\multicolumn{1}{c}{\textbf{\makecell[c]{Parallelizable}}}  &\multicolumn{1}{c}{\bf Architecture} &\multicolumn{1}{c}{\bf Parameters}  &\multicolumn{1}{c}{\bf Accuracy }   \\
\midrule
\multirow{15}{*}{\makecell[c]{S-MNIST / PS-MNIST}}& \multirow{15}{*}{784} & LIF \cite{zhang2023tclif} &\ding{55}  & Feedforward & 85.1k & 72.06\% / 10.00\%   \\
 && ALIF \cite{ALIF} &\ding{55} & Recurrent & 156.3k & 98.70\% / 94.30\% \\ 
&&TC-LIF \cite{zhang2023tclif} &\ding{55} & Recurrent & 155.1k & 99.20\% / 95.36\% \\ 
&& {BHRF \cite{higuchi2024balanced}} &\ding{55} & Recurrent & 68.9k & 99.10\% / 95.20\% \\
&& {DH-LIF \cite{zheng2024temporal}} &\ding{55} & Recurrent & 80.0k & 98.90\% / 94.52\% \\
\cmidrule{3-7}
 &&SPSN \cite{fang2023parallel}*  & \ding{51} & Feedforward & 52.4k & 97.20\% / 82.84\%\\
&&masked PSN \cite{fang2023parallel}*  & \ding{51} & Feedforward &  153.7k & 97.76\% / 97.53\%  \\
 &&PSN \cite{fang2023parallel}* & \ding{51} & Feedforward & 2.5M & 97.90\% / 97.76\% \\
&& {SpikingTCN \cite{ma2025spiking}} & \ding{51} & Convolution & 90.0k &  \quad N.A. / 93.76\% \\
&& {Spike-driven Transformer \cite{yao2023spike}} & \ding{51} & Transformer & 90.0k &  \quad N.A. / 96.21\% \\ 
&& \xy{QKFormer \cite{zhou2024qkformer}} & \ding{51} & \xy{Transformer} & \xy{N.A.} & \xy{ 94.25\% / \ \ N.A. } \\ 
&& {Binary S4D \cite{stan2024learning}} & \ding{51} & State-Space Model & 68.9k & 99.10\% / \ \  N.A. \quad \\
&& {GSU \cite{stan2024learning}} & \ding{51} & State-Space Model & 85.5k & 99.40\% / \ \  N.A. \quad \\
&& {S5-RF \cite{huber2024scaling}} & \ding{51} & State-Space Model & 36.4k & 98.89\% / 95.29\% \\
 &&\multirow{2}{*}{\textbf{PMSN (Ours)}}  &\multirow{2}{*}{\ding{51}}   &\multirow{2}{*}{\textbf{Feedforward}}  & \textbf{66.3k}  & \textbf{99.40\%} / {\textbf{97.51\%}} \\
   & & &&  & \textbf{156.4k}  &  \textbf{99.53\%} / \textbf{97.78\%}  \\
 \midrule
 
 \multirow{19}{*}{SHD}  & \multirow{19}{*}{250} 
& Adaptive axonal delay \cite{sun2023adaptive} &\ding{55} & Feedforward & 109.1k & 92.45\% \\
&& AdLIF+ \cite{deckers2024co} &\ding{55} & Feedforward & 38.7k & 94.19\% \\
&& {TSkip \cite{malettira2025tskips}} & \ding{55} & Feedforward &1.3M &94.71\%\\
&& ALIF \cite{ALIF} & \ding{55} & Recurrent & 141.3k  & 84.40\% \\
&& RadLIF \cite{adLIF}& \ding{55} & Recurrent & 3.9M & 94.62\% \\
&& TC-LIF \cite{zhang2023tclif} & \ding{55} & Recurrent & 141.8k & 88.91\% \\
&& {BHRF \cite{higuchi2024balanced}} & \ding{55} & Recurrent & 108.8k & 92.70\% \\
&& {DH-LIF \cite{zheng2024temporal}} & \ding{55} & Recurrent & 50.0k & 91.34\% \\
\cmidrule{3-7}
&& SPSN \cite{fang2023parallel}* & \ding{51} & Feedforward & 107.1k  & 82.51\% \\
&& masked PSN \cite{fang2023parallel}* & \ding{51} & Feedforward & 122.5k  & 86.00\% \\
&& PSN \cite{fang2023parallel}*  & \ding{51} & Feedforward & 232.5k & 89.75\% \\
&& {DCLS-Delays \cite{hammouamri2024learning}}  & \ding{51} & Convolution & 200.0k & 95.07\% \\ 
&& {\xy{Spikformer \cite{zhouspikformer}}} & \xy{\ding{51}} & \xy{Transformer} & \xy{N.A.} & \xy{85.10\%} \\ 
&& {Spikingformer \cite{zhou2023spikingformer}} & \ding{51} & Transformer & 2.0M & 82.68\% \\ 
&& \xy{TIM \cite{shen2024tim}} & \ding{51} & \xy{Transformer} & \xy{2.6M} & \xy{86.30\%} \\
&& \xy{QKFormer \cite{zhou2024qkformer}} & \ding{51} & \xy{Transformer} & \xy{N.A.} & \xy{88.56\%} \\ 
&& {S5-RF \cite{huber2024scaling}} & \ding{51} & State-Space Model & 214.5k & 91.86\% \\ 
 &&  \multirow{2}{*}{\textbf{PMSN (Ours)}}   &\multirow{2}{*} {\ding{51}} &\multirow{2}{*} {\textbf{Feedforward}}  & \textbf{120.3k} & \textbf{94.25}\%\\
 && &  &  &  \textbf{199.3k} & \textbf{95.10}\%\\
\bottomrule
 \multicolumn{7}{l}{* \hspace{5mm} Our reproduced results based on publicly available codebases
 \hspace{6.2mm} N.A. \hspace{1mm} These results are not publicly available}\\ 
\end{tabular}}
\end{center}
\end{table*}

\section{Effective Temporal Gradient Propagation}
\label{gradient_update}

\xy{
In this section, we provide a theoretical analysis of how the proposed PMSN supports effective multiscale temporal processing. We first derive the gradient flow of PMSN and then discuss how the initialization of compartmental parameters stabilizes the neuronal dynamics and enables multiscale temporal dependency modeling.}

To overcome the gradient discontinuity introduced by the Heaviside and floor functions in the proposed reset operation, we employ surrogate gradient methods \cite{deng2022temporal, bengio2013estimating} to facilitate gradient backpropagation in Eq. \eqref{eq: output_parallel}, yielding a well-defined gradient flow for the PMSN model:
\begin{equation}
\begin{small}
\begin{aligned}
\label{eq:gradw}
\Delta \mathcal{W}^{l}& \propto \frac{\partial \mathcal{L}}{\partial \mathcal{W}^{l}}= \sum_{t=1}^{T}{\frac{\partial \mathcal{L} }{\partial I^{l}[t]}S^{l-1}[t]}, \ \Delta b^{l} \propto \frac{\partial \mathcal{L} }{\partial b^{l}}= \sum_{t=1}^{T}{\frac{\partial \mathcal{L} }{\partial I^{l}[t]}}, \\
\frac{\partial \mathcal{L}}{\partial I^{l}[t]} &=\sum^T_{i=t}\frac{\partial \mathcal{L}}{\partial S^{l}[i]}\frac{\partial S^{l}[i]}{\partial v_s^l[i]}\frac{\partial v_s^l[i]}{\partial I^{l}[t]}+
\sum^{T-1}_{i=t}\frac{\partial \mathcal{L}}{\partial v_s^{l}[i+1]}\frac{\partial v_s^{l}[i+1]}{\partial v_s^{l}[i]}\frac{\partial v_s^l[i]}{\partial I^{l}[t]}\\
&=\underbrace{\frac{\partial \mathcal{L}}{\partial S^{l}[t]}g'[t]\gamma_n}_{\text{Spatial}}+
\underbrace{\sum^T_{i=t}\frac{\partial \mathcal{L}}{\partial S^{l}[i]}g'[i]\Phi_s\Bar{\mathcal{T}}^{i-t}\Phi_c}_{\text{Temporal}},
\end{aligned}\end{small}\end{equation}    
where $g'[t]=\frac{\partial S^l[t]}{\partial v_s^l[t]}$ is the surrogate gradient function, $\mathcal{L}$ is the loss function, and $\mathcal{W}^l$, $b^l$ refer to weight and bias terms of layer $l$, respectively. 
{The first term represents the spatial credit assignment between consecutive network layers, where the gradient propagates from deeper to shallower layers to facilitate hierarchical learning. The second term corresponds to the temporal credit assignment, where gradients flow backward from later to earlier timesteps, facilitating the learning of temporal dependencies across time.}

\xy{
The temporal credit assignment in PMSN is governed by the diagonal state transition matrix $\Bar{\mathcal{T}}=\mathrm{diag}(\lambda_1,\ldots,\lambda_{n-1})$ after eigenvalue decomposition. Each complex-valued eigenvalue $\lambda_i$ characterizes the intrinsic temporal dynamics of one compartment. Specifically, the magnitude of $\lambda_i$ determines the decay rate of temporal information and thus controls the effective memory length, while its phase introduces oscillatory dynamics that arise from inter-compartmental interactions and support temporal integration across different timescales. Together, these eigenvalues determine how PMSN preserves and integrates temporal information across different timescales. Therefore, the initialization of compartmental parameters is important, as it determines the initial distribution of these eigenvalues.
}

{
To provide a stable and expressive starting point, we adopt a structured initialization strategy for the compartmental parameters. The membrane leakage terms on the main diagonal of Eq.~\eqref{eq:enhance_h} keep the compartmental dynamics in a stable regime, preventing uncontrolled divergence and supporting long-sequence information retention, as theoretically analyzed in the \textit{Supplementary Materials}. Meanwhile, the coupling terms on the sub- and super-diagonal entries introduce heterogeneous interactions among neighboring compartments, thereby distributing the dynamics across multiple timescales and enabling PMSN to capture multiscale temporal dependencies, as demonstrated in Fig.~\ref{fig:main_dynamics}.
}

\section{Experimental Results}
In this section, we evaluate the proposed PMSN model with a focus on its multiscale temporal processing capacity, static image classification accuracy, simulation acceleration, and computation efficiency. Unless otherwise stated, all tested PMSNs have $n=5$ compartments to ensure a comparable computational cost to the state-of-the-art (SOTA) parallel spiking neuron model (i.e., 32-receptive-field sliding parallel spiking neuron (SPSN) \cite{fang2023parallel}). We also provide ablation studies in Section \ref{sec:abl} as well as the detailed experimental settings in \textit{Supplementary Materials}.

\subsection{{Temporal Dependency Modeling across Distinct Timescales}}

\begin{table*}[!ht]
    \renewcommand{\arraystretch}{1.2}
    \centering
    \caption{Comparison of different models in handling long-term temporal dependencies on Sequential CIFAR-10 and CIFAR-100 datasets.}
    \resizebox{0.95\textwidth}{!}{
    \begin{tabular}{ccccccccc}
    \hline
         \textbf{Tasks} & \textbf{Timesteps} & \textbf{PMSN} &  \textbf{PMSN (w/o reset)} & \textbf{PSN} & \textbf{masked PSN} & \textbf{SPSN} & \textbf{LIF} & \textbf{LIF (w/o reset)} \\
         \hline
         \makecell[c]{Sequential CIFAR-10 } & \multirow{3}{*}{32} & \textbf{90.97\%} & 89.27\% & 88.45\%& 85.81\%& 86.70\%& 81.50\% &79.50\%\\
         \makecell[c]{Sequential CIFAR-100 } & & \textbf{66.08\%} & 60.82\% & 62.21\% &60.69\% &62.11\% &55.45\%  &53.33\%\\
         No. of Parameters & & {0.54M}  & 0.54M & 0.52M & 0.52M & 0.51M & 0.51M  &0.51M\\ 
         \hline
         \makecell[c]{Sequential CIFAR-10 }& \multirow{2}{*}{1024}& \textbf{82.14\%} & 79.63\% & 55.24\%& 57.83\%& 70.23\%& {45.07\%} &43.30\%\\
         No. of Parameters & & {0.21M} & 0.21M  & 6.47M & 0.38M & 0.18M & 0.18M & 0.18M\\
         \hline
    \end{tabular}}
    \label{tab:row}
\end{table*}

\begin{figure*}[!ht]
\centering
	\includegraphics[width=0.95\textwidth]{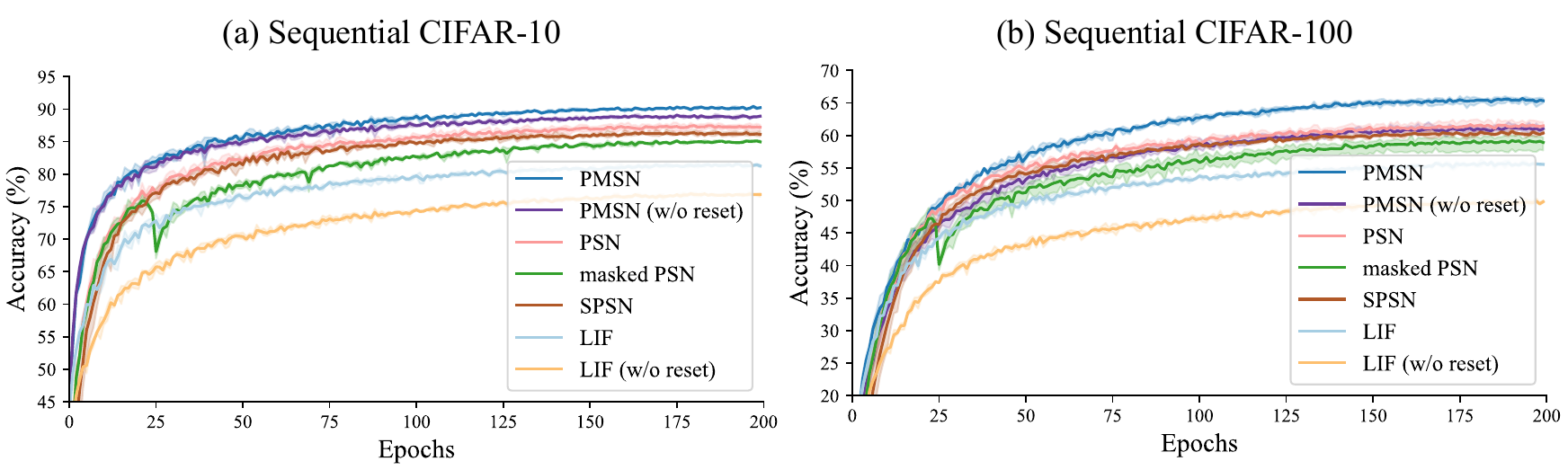}
\caption{The learning curves of PMSN models (with and w/o reset), PSN models, and LIF models (with and w/o reset) on (a) Sequential CIFAR-10 and (b) Sequential CIFAR-100 tasks. \xy{Solid lines and shaded regions denote the mean and standard deviation over three runs with different random seeds.}}
\label{fig: curve}
\end{figure*}

We first compare our PMSN model against other SOTA models on temporal processing tasks involving long-term temporal dependencies. {Our experiments begin with three widely-used neuromorphic benchmarks that are commonly adopted to evaluate temporal dependency modeling}, including Sequential MNIST (S-MNIST) and Permuted Sequential MNIST (PS-MNIST) datasets with $784$ time steps \cite{smnist}, and Spiking Heidelberg Digits (SHD) spoken digit classification dataset with $250$ time steps \cite{cramer2020heidelberg}. 
As summarized in Table~\ref{tab: mnist}, PMSN model achieves the highest accuracy across all spiking neuron models with fewer or comparable amounts of parameters, demonstrating a superior capacity to capture temporal dependencies. 
\xy{Moreover, PMSN consistently achieves comparable or superior performance to advanced spiking architectures specifically designed for temporal processing, including temporal convolutions \cite{ma2025spiking,hammouamri2024learning}, spike-based transformers \cite{yao2023spike,zhou2024qkformer, zhouspikformer, zhou2023spikingformer,shen2024tim}, and SSMs \cite{stan2024learning,huber2024scaling}. For the spike-based transformer baselines, we report publicly available results from prior works~\cite{ma2025spiking,shen2026teformer,shen2024tim,malettira2025tskips}. These findings suggest that detailed modeling of neuronal structural complexity is an important and complementary dimension to architectural innovations in neuromorphic temporal processing.}

We further evaluate our model's ability to establish  spatiotemporal and extended long-term temporal dependencies on the more challenging Sequential CIFAR-10 and CIFAR-100 tasks. For Sequential CIFAR-10, we explore two configurations: column-by-column scanning as per~\cite{fang2023parallel} ($T=32$), which evaluates the model's capacity in integrating both spatial and temporal information, and pixel-by-pixel scanning ($T=1024$), which poses a greater challenge to learning long-term dependency. For Sequential CIFAR-100, we use the column-by-column configuration. To ensure a fair comparison, we employ the same network architecture for each individual task. As shown in Table \ref{tab:row}, our PMSN surpasses the SOTA models by at least 2\% accuracy, showcasing its superiority in multiscale temporal processing. As shown in Fig. \ref{fig: curve}, the learning curves of the PMSN model exhibit faster and more stable training convergence, {aligning well with our theoretical analysis in Section \ref{gradient_update}.}

\begin{table*}[t]
\caption{{Comparison of time-series classification accuracies on the UEA datasets. Test accuracy is averaged over five seeds. The best score is highlighted in \textbf{bold} and the second is \underline{underlined}.}}
\label{tab: uea}
\centering
\small
{
\begin{tabular}{lcccccccc}
\toprule
\textbf{Task}  & \multirow{2}{*}{\textbf{SNN}} & \textbf{Worms} & \textbf{SCP1} & \textbf{SCP2} & \textbf{Ethanol} & \textbf{Heartbeat} & \textbf{Motor} & \multirow{2}{*}{\textbf{Avg.}} \\
Seq. Length & & (17,984) & (896) & (1,152) & (1,751) & (405) & (3,000) & \\
\midrule
NRDE \cite{morrill2021neural}  & \ding{55} & 83.9 & 80.9 & 53.7 & 25.3 & 72.9 & 47.0 & 60.6 \\
Log-NCDE \cite{walker2024log} & \ding{55} & 85.6 & 83.1 & 53.7 & \underline{34.4} & 75.2 & 53.7 & 64.3 \\
LRU \cite{orvieto2023resurrecting}     & \ding{55} & 87.8 & 82.6 & 51.2 & 21.5 & \underline{78.4} & 48.4 & 61.7 \\
S5  \cite{s5}      & \ding{55} & 81.1 & \textbf{89.9} & 50.5 & 24.1 & 77.7 & 47.7 & 61.8 \\
S6  \cite{gu2024mamba}    & \ding{55} & 85.0 & 83.2 & 49.9 & 26.4 & 76.5 & 51.3 & 62.0 \\
Mamba \cite{gu2024mamba}    & \ding{55} & 70.9 & 80.7 & 48.2 & 27.9 & 76.2 & 47.7 & 58.6 \\
LinOSS-IMEX \cite{ruschoscillatory}& \ding{55} & 80.0 & 87.5 & \textbf{58.9} & 29.9 & 75.5 & \underline{57.9} & 65.0 \\
LinOSS-IM  \cite{ruschoscillatory} & \ding{55} & \textbf{95.0} & 87.8 & 58.2 & 29.9 & 75.8 & \textbf{60.0} & \underline{67.8} \\
\midrule
\textbf{PMSN (Ours)}& \ding{51} & \textbf{95.0} & \underline{88.7} & \textbf{58.9} & \textbf{34.9} & \textbf{78.7} & 56.1 & \textbf{68.7} \\
\bottomrule
\end{tabular}}
\end{table*}

Furthermore, we would like to stress the importance of the neuronal reset mechanism that has been neglected by many previous works \cite{fang2023parallel}.
As presented in Table \ref{tab:row}, the accuracy consistently improves for both PMSN and LIF models after incorporating the reset mechanism. 
This is because the reset mechanism prevents the membrane potential from becoming excessively high, yielding a smoother distribution of membrane potentials across time that can facilitate stable information flow.

{Beyond these vision-based and audio processing benchmarks, we further assess the scalability and practicality of PMSN on the more challenging UEA multivariate time-series classification benchmark \cite{walker2024log} and the PPG-DaLiA time-series regression benchmark \cite{reiss2019deep}. Both benchmarks involve significantly longer sequences, with several tasks containing more than 10,000 timesteps, and capture a broad range of real-world temporal processing scenarios, ranging from scientific time-series analysis to practical applications in healthcare. Following prior work, we adopt the same preprocessing and evaluation protocols to ensure a fair comparison \cite{walker2024log,ruschoscillatory}. As reported in Tables~\ref{tab: uea} and~\ref{tab: ppg}, PMSN achieves competitive or superior performance compared with advanced non-spiking architectures that are widely used for temporal processing, including linear RNNs and SSMs, while being the only spike-based model among the compared methods. These results demonstrate that PMSN scales effectively to very long and challenging sequential tasks, underscoring its applicability to real-world temporal processing scenarios and its competitiveness beyond the spiking domain.}

\begin{table}[t]
\centering
\caption{{Comparison of regression errors on the PPG-DaLiA dataset with a sequence length of 49,920, averaged over five different seeds. The best score is highlighted in \textbf{bold}.}}
\label{tab: ppg}
\small
{
\begin{tabular}{lcc}
\toprule
\textbf{Model} & \textbf{SNN} & \textbf{MSE $\times 10^{-2}$} \\
\midrule
NRDE \cite{morrill2021neural}  & \ding{55} & 9.90 $\pm$ 0.97\\
Log-NCDE \cite{walker2024log} & \ding{55} & 9.56 $\pm$ 0.59 \\
LRU \cite{orvieto2023resurrecting} & \ding{55} & 12.17 $\pm$ 0.49 \\
S5 \cite{s5} & \ding{55} & 12.63 $\pm$ 1.25\\
S6 \cite{gu2024mamba} & \ding{55} & 12.88 $\pm$ 2.05 \\
Mamba \cite{gu2024mamba}& \ding{55} & 10.65 $\pm$ 2.20 \\
LinOSS-IMEX \cite{ruschoscillatory}  & \ding{55} & 7.50 $\pm$ 0.46 \\
LinOSS-IM \cite{ruschoscillatory} & \ding{55} & 6.40 $\pm$ 0.23 \\
\midrule
\textbf{PMSN (Ours)} & \ding{51} & \textbf{5.17 $\pm$ 0.53}  \\
\bottomrule
\end{tabular}}
\end{table}

\subsection{{Scalability on Large-scale Image Classification}}
{While the preceding evaluations primarily focus on temporal processing benchmarks, it is also important to examine whether the proposed spiking neuron model can scale to larger network architectures and datasets. To assess this scalability, we integrate PMSN into ResNet-18 and ResNet-34 architectures and evaluate its performance on the ImageNet-1K benchmark \cite{deng2009imagenet}, with the results summarized in Table~\ref{tab:imagenet}.
The results show that PMSN achieves SOTA accuracy against other spiking models with identical architectures, reaching 70.98\% on ImageNet-1K, a 3\% absolute improvement over the IF baseline. This consistent improvement across both sequential benchmarks and large-scale image recognition tasks demonstrates the effectiveness and scalability of the proposed neuron model.}

\begin{table}[t]
\renewcommand{\arraystretch}{1.15}
\centering
\caption{{Comparison on the ImageNet-1K image classification dataset.}}
\label{tab:imagenet}
\resizebox{0.49\textwidth}{!}{
\begin{tabular}{lcccc}
\toprule
\textbf{Approach} & \textbf{Architecture} & \textbf{Param.} & \textbf{Timesteps} &\textbf{Acc.} \\ 
\midrule
{Spikformer \cite{zhouspikformer}} & Spikformer-8-384 & \multirow{3}{*}{17M} & 4 &70.24\% \\
{Spike-driven Transformer \cite{yao2023spike}} & S-Transformer-8-384 &  & 4 & 72.28\%  \\
{Spikingformer \cite{zhou2023spikingformer}} & Spikingformer-8-384 & & 4 & \textbf{72.45\%} \\
\midrule
MPBN \cite{mpbn} & ResNet-18 & \multirow{9}{*}{12M}  & 4 & 63.14\% \\
{InfLoR-SNN \cite{guo2023inflor}}  & ResNet-18 & & 4 & 64.78\% \\
{IF \cite{fang2021deep}} & SEW ResNet-18 & & 4 & 63.18\% \\
{GLIF \cite{GLIF}} & MS-ResNet-18  &  & 6 & 68.11\% \\ 
{Attention SNN \cite{yao2023attention}} & MS-ResNet-18 & &  1 & 63.97\% \\
{PSN+TET \cite{fang2023parallel}} & SEW ResNet-18 & & 4 & 67.63\% \\ 
{IMP+LTS \cite{shen2024rethinking}} & SEW ResNet-18 & & 4 & 65.38\% \\
{SMA-ResNet \cite{shan2025advancing}} & SMA-MS-ResNet-18 &  & 6 & 68.46\% \\
{\textbf{PMSN (Ours)}} &  \textbf{SEW ResNet-18} & &  \textbf{4} & \textbf{68.77\%} \\
\midrule
MPBN \cite{mpbn} & ResNet-34 & \multirow{12}{*}{22M} & 4 & 64.71\% \\
{InfLoR-SNN \cite{guo2023inflor}}  & ResNet-34 & & 4 & 65.54\% \\ 
{IF \cite{fang2021deep}} & SEW ResNet-34 & & 4 & 67.04\% \\
{GLIF \cite{GLIF}} & ResNet-34  &  & 4 & 67.52\% \\ 
{TET \cite{deng2022temporal}}& SEW ResNet-34 & & 4 & 68.00\% \\
{Attention SNN \cite{yao2023attention}} & MS-ResNet-34 && 1 & 69.15\% \\
{PSN+TET \cite{fang2023parallel}} & SEW ResNet-34 & & 4 & 70.54\% \\
{IMP+LTS \cite{shen2024rethinking}} & SEW ResNet-34 & & 4 & 68.90\% \\
{SMA-ResNet \cite{shan2025advancing}} & SMA-MS-ResNet-34 &  & 6 & 70.19\% \\
\multirow{2}{*}{{ANN-SNN Distillation \cite{yang2025efficient}}} & SEW ResNet-34 & & 4 & 68.12\% \\
& ResNet-34 & & 4 & 70.64\% \\
{\textbf{PMSN (Ours)}} &  \textbf{SEW ResNet-34} & &  \textbf{4} & \textbf{70.98\%} \\
\bottomrule
\end{tabular}}
\end{table}

{At the same time, it is worth noting that recent transformer-based architectures introduce innovations tailored for static image recognition and can achieve strong ImageNet performance \cite{zhouspikformer,yao2023spike,zhou2023spikingformer}. Under comparable experimental settings, PMSN attains comparable performance to these approaches, despite relying solely on convolutional backbones and being primarily designed for temporal processing rather than static image feature extraction.}

\begin{figure*}[t]
    \centering
\includegraphics[scale=0.58,trim= 0 1022 0 20, clip]{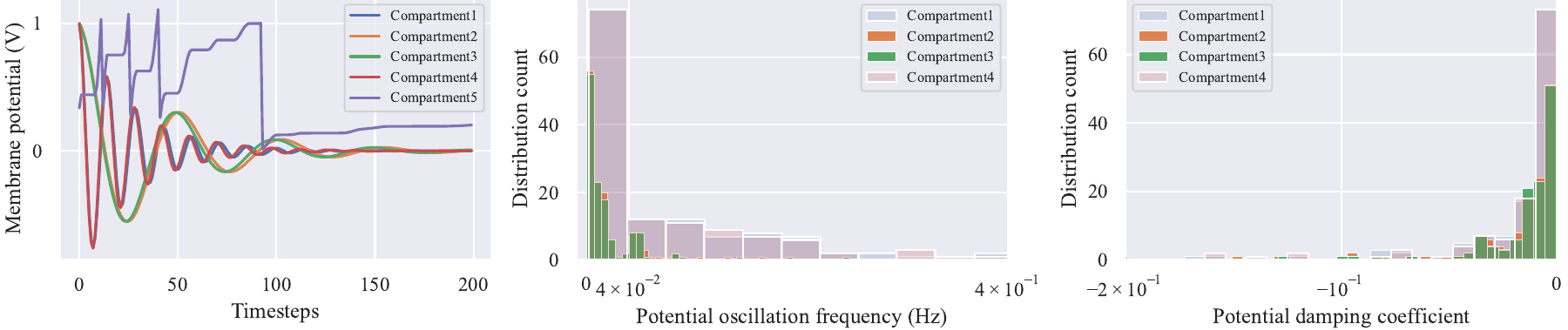}
\caption{{Visualization of PMSN model dynamics ($n=5$). \textbf{Left:} The impulse response of different compartments within one PMSN neuron, indicating the multi-timescale properties of a single PMSN neuron. Each hidden compartment has a dynamic coefficient $\lambda_i=e^{a+\boldsymbol{i}b}$ and exhibits damped oscillations in response to inputs, while the compartment$5$ handles spike generation and reset.}
\textbf{Middle:} The distribution of oscillation frequencies {$b/2\pi$}, and \textbf{Right:} damping coefficients {$a$} across neurons in the same layer, revealing heterogeneous temporal dynamics across timescales.
}
\label{fig:main_dynamics} 
\end{figure*}
\subsection{Visualization of Multi-compartment Dynamics}

Having demonstrated the superior performance of PMSN, we next analyze how PMSN integrates multiscale temporal information through compartmental interactions. We first visualize the single-neuron impulse response in Fig.~ \ref{fig:main_dynamics}. After receiving an impulse at $t=0$, different compartments exhibit damped oscillations with distinct frequencies and decay rates, characterized by $\lambda_i=e^{a+ib}$. This reflects the system's multiscale properties: compartments with higher oscillation frequencies correlate with faster timescales, while lower frequencies indicate slower scales \cite{jaeger2021dimensions}. 
This compartmental synergy enables the integration of information across various frequency domains and time spans, facilitating multiscale temporal processing.

{We further examine the distribution of oscillation frequencies and damping coefficients across neurons in one layer. As shown in the middle and right panels of Fig.~\ref{fig:main_dynamics}, PMSNs within the same population exhibit heterogeneous temporal dynamics, allowing the network to cover a broader spectrum of timescales.}

\begin{figure*}[t]
\centering
\includegraphics[width=0.8\textwidth,trim= 0 0 0 10, clip]{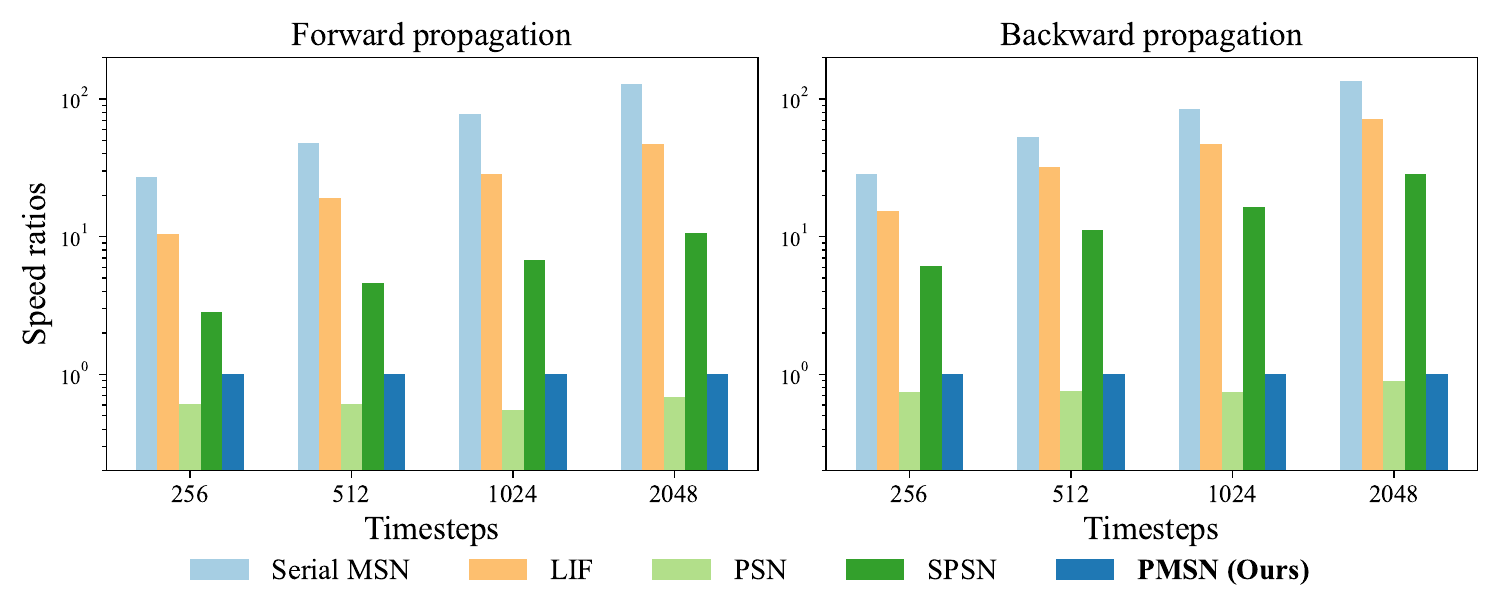}
\caption{Comparison of speed ratios $t_i/{t_{PMSN}}$ among evaluated neuron models, {where $t_i$ is the wall-clock runtime per propagation for model $i$. }} 
\label{fig: speedup}
\end{figure*}

\subsection{Simulation Acceleration}
To quantitatively evaluate the proposed acceleration techniques, we record the wall-clock forward- and backward-propagation times of PMSN using the GPU-enabled PyTorch library. Besides that, we also compare them against a range of serial and parallel models, including a serial MSN with neuronal dynamics identical to those of PMSN, LIF, PSN, and SPSN. Fig. \ref{fig: speedup} shows the acceleration ratios across diverse sequence lengths {under the same single-hidden layer network architecture consisting of 256 neurons.}
Our parallel scheme shows a substantial speed-up over serial models, achieving training speed-up ratios up to $134 \times$ and $71 \times$ compared to the serial MSN and LIF models, respectively. When compared with SOTA parallel models, our PMSN outperforms SPSN but is slower than PSN. This can be attributed to a larger number of neuronal compartments used (i.e., five in PMSN vs. one in PSN) and less efficient implementation of FFT operations compared to PSN's matrix multiplication in the current GPU acceleration framework. Nevertheless, this slight slowdown is worthwhile considering the overall effectiveness of our model. Additionally, we observe a positive correlation between the speed-up ratio and sequence length, indicating greater relative acceleration for longer sequences.

\subsection{Theoretical Computational Cost Analysis}

\label{sec:cost}
\begin{figure*}[t]
\centering
\includegraphics[width=0.8\textwidth,trim= 10 0 0 10, clip]{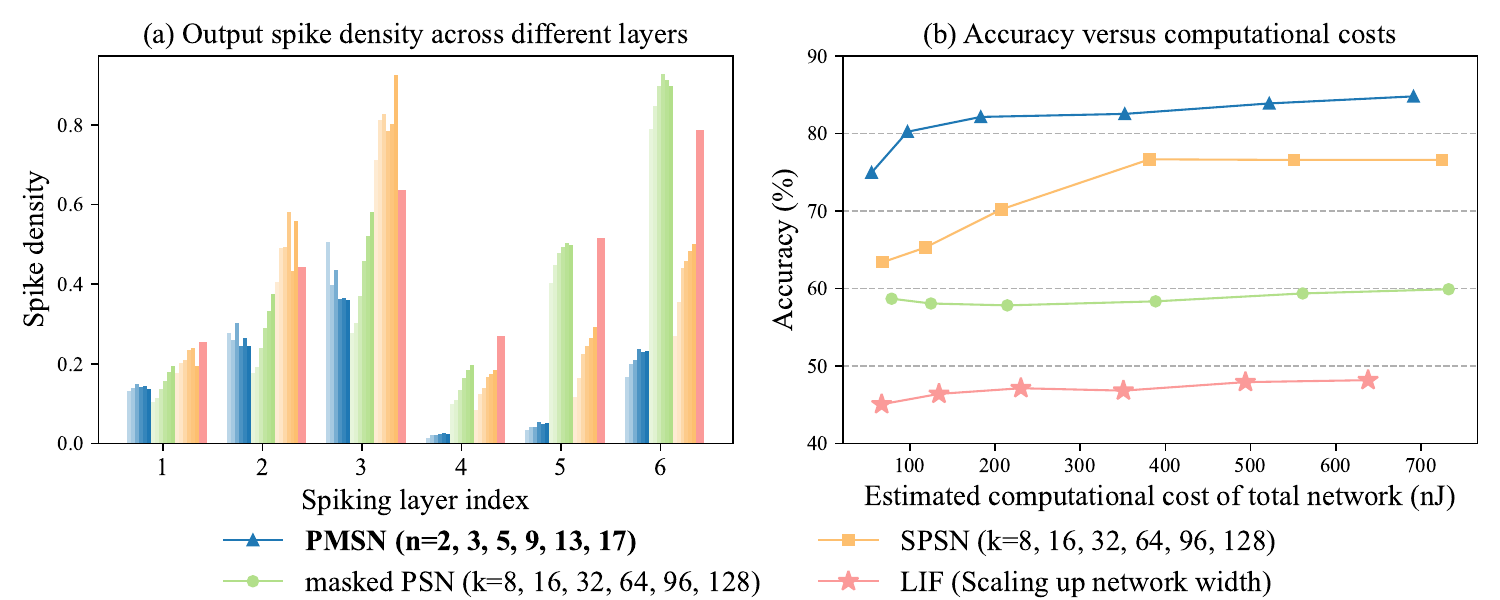}\hspace{0em}
\caption{Spike sparsity, accuracy, and estimated computational cost across neuron models and model sizes; darker bars in (a) indicate larger models.}
\label{fig: cost_curve}
\end{figure*}

{The PMSN offers an energy-efficient, hardware-friendly solution for practical applications. To systematically characterize these properties, we first analyze the spike sparsity and theoretical computational cost of PMSN in comparison with the PSN family \cite{fang2023parallel} and conventional LIF neurons. 
\xy{All experiments are conducted on the pixel-by-pixel Sequential CIFAR-10 task using identical backbone architectures to ensure a fair comparison. To compare models across different scales, we vary the compartment number $n$ for PMSN, the temporal receptive field size $k$ for masked PSN and SPSN, and the network size for LIF neurons. The total cost is computed by considering both synaptic operations and neuronal operations.} As shown in Fig.~\ref{fig: cost_curve}(a), PMSN exhibits the lowest spike density among all compared models. This efficiency arises from two key mechanisms: firstly, the mutual inhibitory effect resulting from the coupling compartments may contribute to the spike frequency adaptation. Moreover, richer compartmental dynamics reduce the spike-based computational workload and suppress spike redundancy.}

{To estimate the theoretical computational cost, we count the number of Multiply–Accumulate (MAC) and Accumulate (AC) operations per inference. 
The analytical cost of each neuron model is provided in \textit{Supplementary Materials}. Using CMOS energy estimates at 45 nm \cite{CMOS} ($E_{AC}=0.9$ pJ, $E_{MAC}=4.6$ pJ), Fig.~\ref{fig: cost_curve}(b) shows that PMSN consistently achieves higher accuracy than other spiking neuron models at comparable computational costs.

\subsection{{Hardware Implementation and Energy Efficiency}}

\xy{To evaluate the practical deployability and efficiency of PMSN, we conduct an FPGA-based neuromorphic hardware implementation} and compare its energy efficiency against conventional ANN implementations as well as a baseline SNN with LIF neurons. Details of the hardware platforms, software environments and the measurement pipeline are provided in \textit{Supplementary Materials}.

\begin{table*}[t]
    \centering
    \caption{Accuracy–efficiency comparison of ANNs on CPU/GPU and SNNs with LIF or PMSN neurons on an FPGA-based neuromorphic hardware.}
    \resizebox{0.98\textwidth}{!}{%
    {
    \begin{tabular}{lccccccccc}
    \toprule
          \textbf{Architecture} &  \textbf{Neuron Model}  & \textbf{Device}& \textbf{Hidden Dimension} & \textbf{Batch Size} & \textbf{Accuracy}  &  \makecell[c]{\textbf{Throughput} \\ \textbf{(Samples/s)}} & \makecell[c]{\textbf{Power} \\ \textbf{(W)}} &  \makecell[c]{\textbf{Energy Efficiency}\\ \textbf{(Samples/J)}}\\
    \midrule
           \multirow{4}{*}{Feedforward ANN}&  \multirow{4}{*}{ReLU Activation} & \multirow{2}{*}{CPU} & 64-256-256 & 32 & -  & 28.33 & 83.87 & 0.34  \\
          &   & & 64-256-256 & 64 & -  & 55.28 & 88.26 & 0.63  \\
          \cmidrule{3-9}
          &  & \multirow{2}{*}{GPU} & 64-256-256 & 32 & - &  45.48  & 20.20 & 2.25  \\
          &  & & 64-256-256 & 64 & -  & 94.90 & 20.32 & 4.67  \\
    \midrule
          Feedforward SNN & LIF & Neuromorphic & 64-256-256 & 1 &  41.47\%  & 25.58  & 0.138 & 185.36 \\
    \midrule
        \multirow{2}{*}{\textbf{Feedforward SNN}} & \multirow{2}{*}{\textbf{PMSN (Ours)}} & \multirow{2}{*}{\textbf{Neuromorphic}} & 64-256-256 & 1 &  \textbf{97.67\%}  &25.59   &0.313  & \textbf{81.76} \\
         & &  & 64-64-64 & 1 &  \textbf{96.02\%}  & 268.97  &0.313  & \textbf{859.33} \\
    \bottomrule    
    \end{tabular}}}
    \label{tab:energy_compare}
\end{table*}

\begin{figure}
    \centering
    \includegraphics[width=0.95\linewidth]{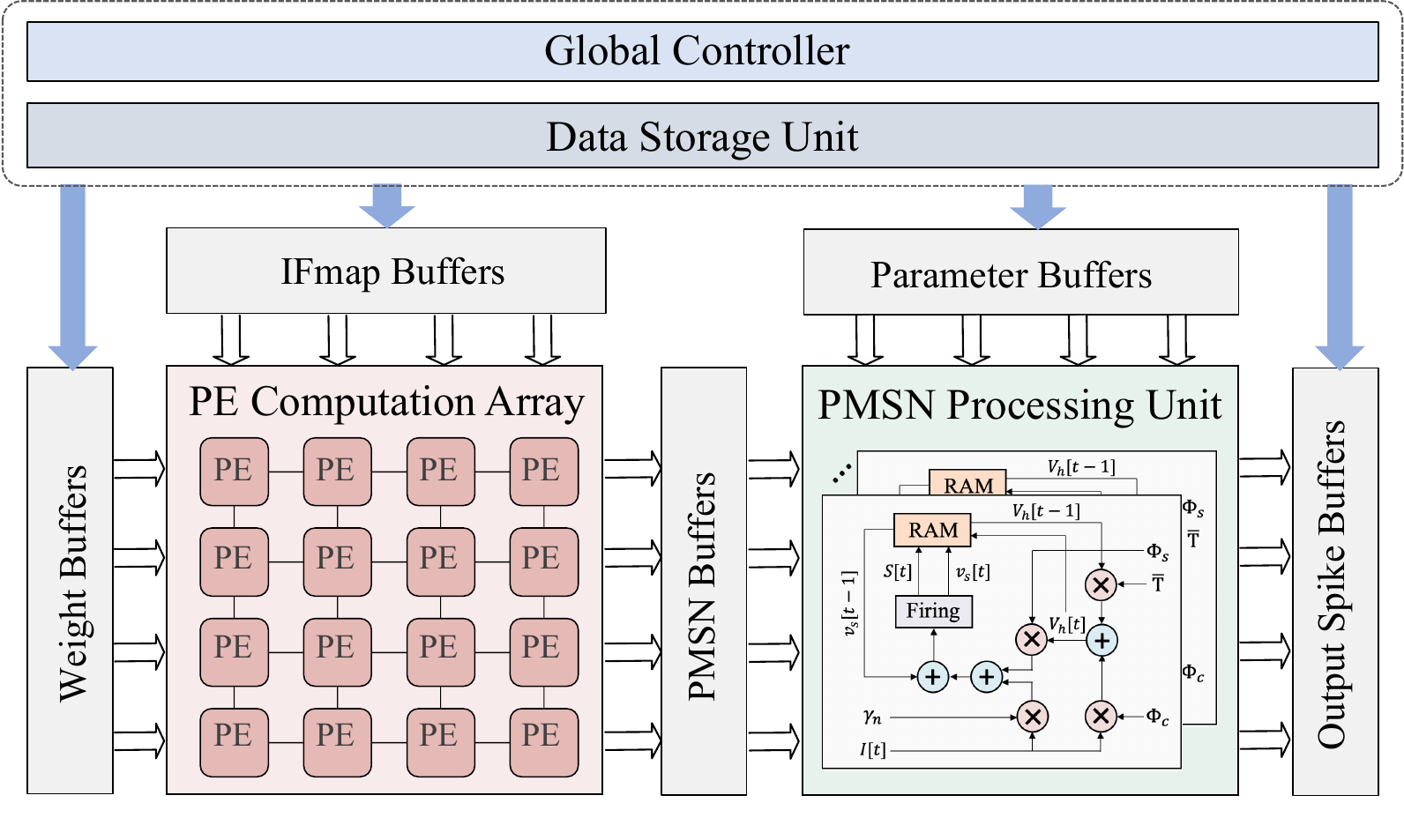}
    \caption{{\textbf{Overall framework of the FPGA-based hardware accelerator tailored for PMSN.}
    The accelerator comprises: \textbf{a global controller}, coordinating the overall execution flow; \textbf{PMSN processing units}, computing the internal dynamics of PMSN neurons, including parallel matrix operations, adder-tree accumulation, membrane potential update, and spike generation; \textbf{a PE computation array}, accelerating linear-layer computations; and \textbf{a data storage unit}, storing synaptic weights and intermediate results. 
    The accelerator adopts a layer-wise execution paradigm, in which the PMSN units and the PE array are shared across layers and reconfigurable, enabling efficient adaptation to diverse network architectures.}
    }
    \label{fig: deployment}
\end{figure}

{Specifically, we develop a dedicated FPGA-based neuromorphic accelerator tailored for PMSN, as illustrated in Fig. \ref{fig: deployment}. This hardware implementation preserves event-driven inter-neuron communication while retaining the rich internal neuronal dynamics of PMSN. Building upon this hardware deployment, we conduct an energy efficiency evaluation of PMSN on the PS-MNIST task and compare it against conventional ANN implementations executed on CPU and GPU platforms, and a baseline feedforward SNN with LIF neurons implemented on the same FPGA platform. The results are summarized in Table~\ref{tab:energy_compare}, where energy efficiency is defined as the ratio between throughput and power consumption.
Compared to ANN implementations on CPU and GPU, PMSN achieves two orders of magnitude improvement in energy efficiency, demonstrating its suitability for real-time temporal signal processing in edge scenarios. Moreover, PMSN exhibits a more favorable accuracy-efficiency trade-off than baseline LIF neurons. In particular, a PMSN configuration with network size comparable to the LIF baseline achieves substantially higher accuracy at the cost of moderately reduced energy efficiency, while a more compact variant achieves markedly higher energy efficiency than the LIF baseline while still delivering significantly higher accuracy.}

\subsection{Ablation Study of Compartment Number}
\label{sec:abl}

\xy{
PMSN provides the flexibility to adjust the number of neuronal compartments according to task complexity and computational resources. To examine this property, we evaluate the effect of compartment number on Sequential CIFAR-10 and CIFAR-100 ($T=32$). As summarized in Table~\ref{tab:ablation}, increasing the number of compartments consistently improves accuracy on both datasets, indicating that larger compartment numbers provide additional temporal modeling capacity.
}

\xy{
However, this performance gain comes at the cost of increased neuronal computation. As shown in Fig.~\ref{fig: cost_curve}(b), increasing $n$ from $2$ to $17$ substantially improves classification accuracy while incurring a more than 10-fold increase in theoretical computational cost, revealing a trade-off between model expressiveness and computational efficiency. 
}

\begin{table}[t]
    \centering
    \caption{Ablation study of compartment number $n$ {on Sequential CIFAR-10 and CIFAR-100 with $T=32$.}}
    \renewcommand{\arraystretch}{1.1}
    \resizebox{0.48\textwidth}{!}{
    \begin{tabular}{ccccccc}
    \hline
         \multirow{2}{*}{\textbf {Dataset}} & \multicolumn{6}{c}{\textbf{Compartment number $\boldsymbol{n}$}}  \\
         \cmidrule{2-7}
         &  2 &  3 & 4 & \textbf{5} & 9 & 17 \\
         \hline
         \makecell[c]{Sequential\\CIFAR-10} & 88.79\% & 90.49\% & 90.75\% & 90.97\% & 91.05\% & 91.43\%\\
         \hline
         \makecell[c]{Sequential\\CIFAR-100} & 61.84\% & 65.16\% & 65.83\% & 66.08\% & 66.51\% & 66.81\% \\
    \hline
    \end{tabular}}
    \label{tab:ablation}
\end{table}

\section{Discussion and Conclusion}
In this work, we proposed a generalized multi-compartment neuron model with superior capacity in multiscale temporal processing. Furthermore, we introduced a parallel implementation for this model, enabling accelerated training on GPU-accelerated ML frameworks. Experimental results demonstrated its strong ability to establish temporal dependencies across multiple timescales, substantially accelerate training, scale to challenging tasks, and achieve a favorable trade-off between accuracy and energy efficiency.

While PMSN has demonstrated accurate and efficient neuromorphic temporal processing, how this neuron-level advance interacts with contemporary Transformer-based architectures remains an interesting question. Transformers provide architecture-level global representation learning, whereas PMSN enhances multiscale temporal integration within individual spiking units. Recent hybrid Transformer--SSM models suggest the benefit of combining global attention with structured temporal dynamics~\cite{rensamba,lenz2025jamba}. Therefore, integrating PMSN with attention-based architectures remains a promising direction for future work.

Moreover, our FPGA-based neuromorphic accelerator evaluation provided hardware-oriented evidence for the potential deployability of PMSN on neuromorphic platforms, positioning it as a promising solution for accurate and energy-efficient real-time temporal processing in edge applications. 
Its compartmental dynamics can also be formulated as programmable state-space neuronal dynamics, suggesting potential compatibility with neuromorphic chips that support programmable neurons \cite{gonzalez2024spinnaker2} and efficient state-space inference \cite{meyer2025diagonal}.


\end{document}